%% file: main.tex
\documentclass[10pt,twocolumn,letterpaper]{article}

\usepackage{cvpr}              


\definecolor{cvprblue}{rgb}{0.21,0.49,0.74}
\usepackage[pagebackref,breaklinks,colorlinks,allcolors=cvprblue]{hyperref}
\usepackage{multirow}                 
\usepackage{multicol}                 
\usepackage{float}                    
\usepackage{makecell}                 
\usepackage{booktabs}                 
\usepackage{graphicx} 
\usepackage[dvipsnames]{xcolor}
\usepackage{colortbl}  
\usepackage{pifont}
\usepackage{amssymb}
\usepackage{xcolor}
\usepackage{marvosym}
\renewcommand{\thefootnote}{}

\def\paperID{4235} 
\def\confName{CVPR}
\def\confYear{2025}

\title{FastVGGT: Training-Free Acceleration of Visual Geometry Transformer
}

\author{You Shen\textsuperscript{1}, Zhipeng Zhang\textsuperscript{2}, Yansong Qu\textsuperscript{1}, Xiawu Zheng\textsuperscript{1}, Jiayi Ji\textsuperscript{1}, Shengchuan Zhang\textsuperscript{1}, Liujuan Cao\textsuperscript{1}\footnotemark[1] \\
\textsuperscript{1}Key Laboratory of Multimedia Trusted Perception and Efficient Computing,\\
Ministry of Education of China, Xiamen University\\
\textsuperscript{2}AutoLab, School of Artiﬁcial Intelligence, Shanghai Jiao Tong University
}

\usepackage{graphicx}
\usepackage{caption}
\usepackage{lipsum}

\begin{document}

\twocolumn[{
\maketitle
\begin{center}
    \captionsetup{type=figure}
    \includegraphics[scale=0.69]{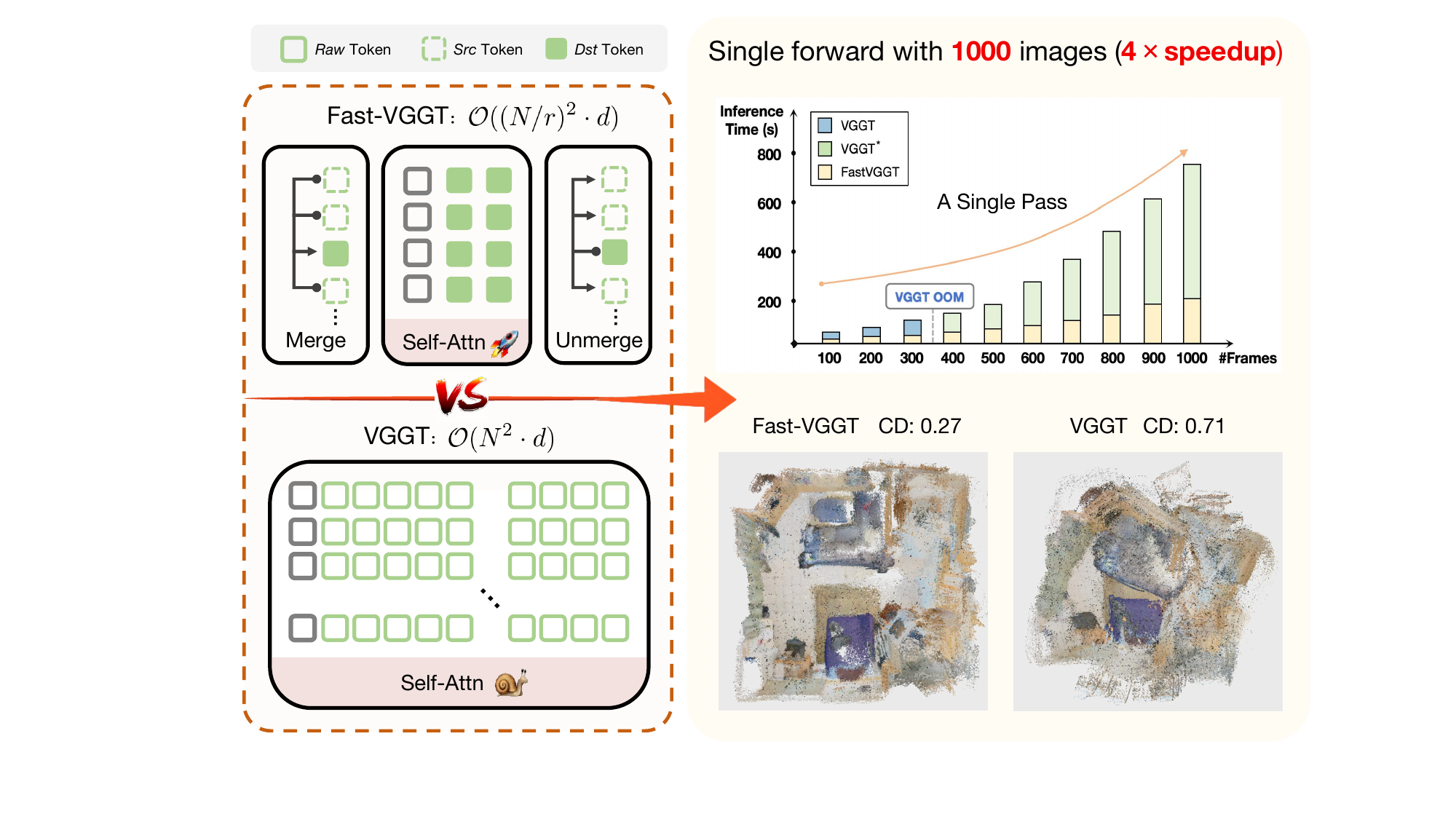}\vspace{0mm}
    \captionof{figure}{We propose FastVGGT, a training-free framework that processes 1,000 images in a single inference, achieving 4$\times$ faster while mitigating error accumulation. VGGT* refers to VRAM-Efficient of VGGT, enabling larger inputs.}\label{P1}\vspace{3mm}
\end{center}
}]

\maketitle
\renewcommand{\thefootnote}{\fnsymbol{footnote}}
\footnotetext[1]{Corresponding Author.}

\input{sec/0_abstract}    
\input{sec/1_intro}

\input{sec/2_related}

\input{sec/3_method}

\input{sec/4_experiment}

{
    \small
    \bibliographystyle{ieeenat_fullname}
    \bibliography{main}
    
}


\end{document}

%% file: sec/0_abstract.tex
\begin{abstract}

Foundation models for 3D vision have recently demonstrated remarkable capabilities in 3D perception. However, scaling these models to long-sequence image inputs remains a significant challenge due to inference-time inefficiency. In this work, we present a detailed analysis of VGGT, a state-of-the-art feed-forward visual geometry model and identify its primary bottleneck. Visualization further reveals a token collapse phenomenon in the attention maps. Motivated by these findings, we explore the potential of token merging in the feed-forward visual geometry model. Owing to the unique architectural and task-specific properties of 3D models, directly applying existing merging techniques proves challenging. To this end, we propose FastVGGT, which, for the first time, leverages token merging in the 3D domain through a training-free mechanism for accelerating VGGT. we devise a unique token partitioning strategy tailored to 3D architectures and tasks, effectively eliminating redundant computation while preserving VGGT’s powerful reconstruction capacity. Extensive experiments on multiple 3D geometry benchmarks validate the effectiveness of our approach. Notably, with 1000 input images, FastVGGT achieves a 4$\times$ speedup over VGGT while mitigating error accumulation in long-sequence scenarios. These findings underscore the potential of token merging as a principled solution for scalable 3D vision systems. Code is available at: https://mystorm16.github.io/fastvggt/.

\end{abstract}

%% file: sec/1_intro.tex
\section{Introduction}
\label{sec:intro}


Inferring the 3D geometric structure of a scene from visual inputs, is critical for enabling machines to understand and interact with the physical world. Recent advances in deep learning have catalyzed a paradigm shift in 3D geometric estimation~\cite{wang2024dust3r,zhang2025flare,qu2023sg,wang2025vggt,qu2025deocc}, enabling a move from iterative, optimization-based pipelines to end-to-end neural networks that directly infer geometry from raw visual inputs. This transformation is exemplified by large-scale architectures like DUSt3R~\cite{wang2024dust3r} and its follow-ups~\cite{yang2025fast3r,tang2025mv,leroy2024grounding}, which showcase a remarkable capacity to reason about complex geometric relationships across image pairs.

Building upon this line of research, VGGT~\cite{wang2025vggt} marks a significant advance. Its transformer-based, feed-forward architecture directly regresses key 3D attributes, including camera parameters, depth maps, and point tracks, to achieve highly stable and accurate reconstructions. While this establishes VGGT as a state-of-the-art framework for 3D scene understanding, its scalability is impeded by two critical bottlenecks. First, the model's reliance on dense global token interactions across views or frames results in prohibitive computational costs. Although used techniques like Flash-Attention mitigate the memory complexity from $O(n^{2})$ to $O(nd)$, the underlying time complexity remains quadratic at $O(n^{2}d)$. Second, the global attention mechanism, essential for capturing inter-frame relations, is susceptible to error accumulation. As the token space expands with each new frame, minor inaccuracies are amplified, leading to significant prediction drift. Collectively, these limitations restrict VGGT's applicability in large-scale scenarios and motivate the development of more efficient and scalable architectures.

\begin{figure}[t]
	\centering
	\includegraphics[scale=0.35]{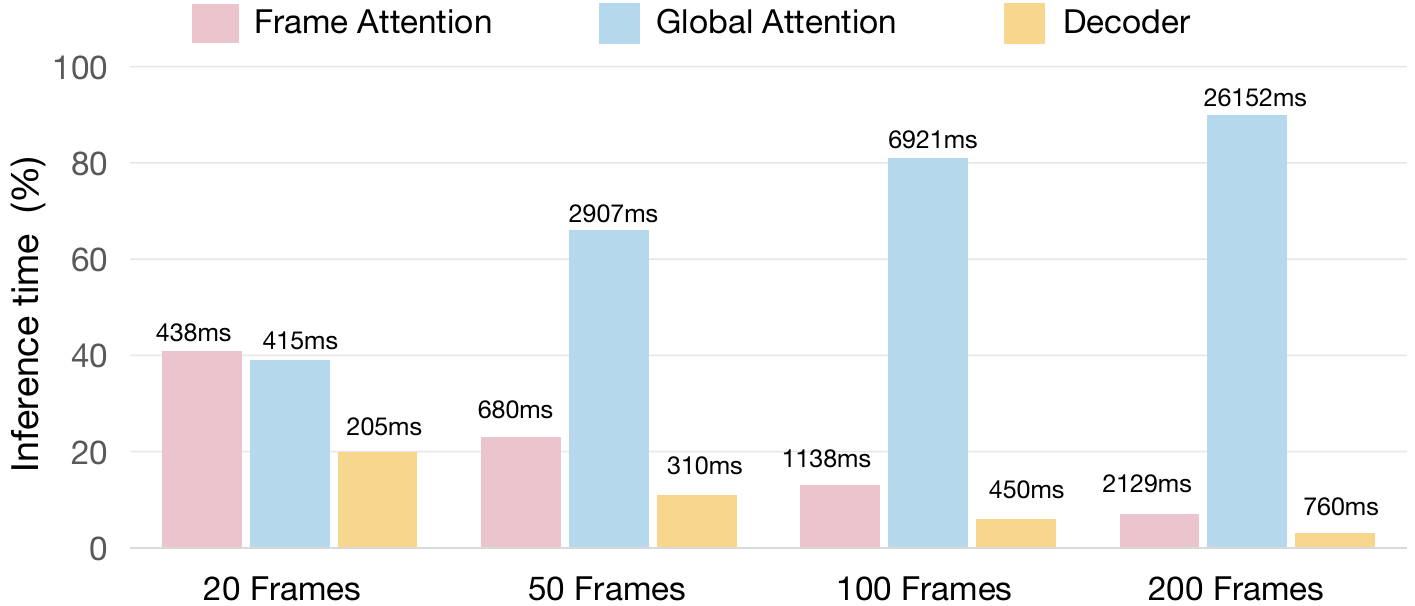}\vspace{0mm}
	\caption{Component-wise analysis of VGGT inference time. As the number of input frames grows, the Global Attention module increasingly dominates the computational cost.
}\label{P2}\vspace{0mm}
\end{figure}

\begin{figure*}[t]
	\centering
	\includegraphics[scale=0.8]{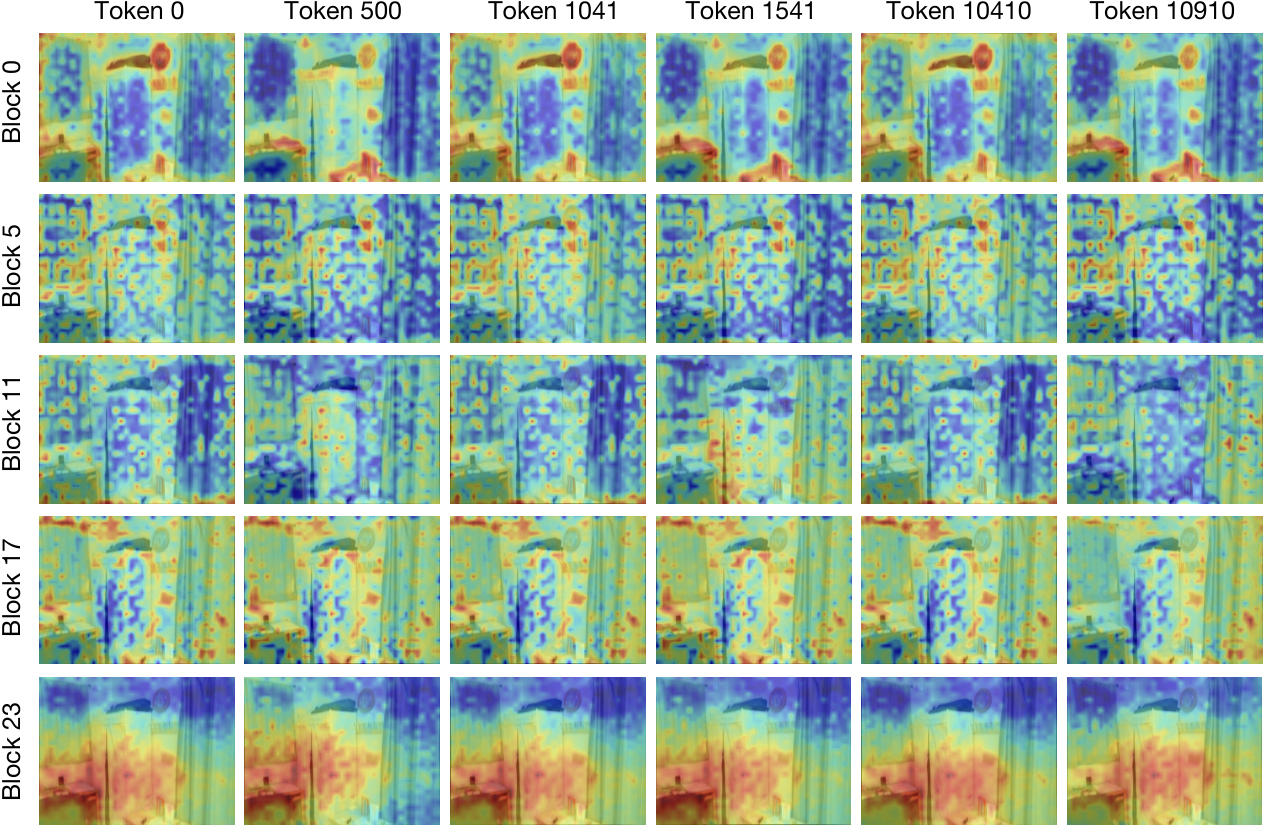}\vspace{0mm}
	\caption{Visualizations of the Global Attention maps in VGGT, using six representative tokens (including the camera token and several image tokens), show that at every stage the attention patterns of different tokens exhibit a strong degree of similarity.}\label{P3}\vspace{-1mm}
\end{figure*}

To pinpoint the primary inference bottlenecks in VGGT, we first conducted a detailed, component-wise performance analysis, as illustrated in Figure~\ref{P2}. The analysis reveals that while the computational costs of ``Frame Attention'' (intra-frame interaction) and ``Global Attention'' (inter-frame interaction) are comparable for short sequences, the cost of Global Attention escalates rapidly with sequence length, eventually dominating the entire runtime profile. This finding motivates our central research question: \textit{Can the computational inefficiency of Global Attention be mitigated without compromising VGGT's capability?} To investigate this possibility, we visualized the attention maps in Figure.~\ref{P3}, which uncovered a crucial insight that attention patterns across tokens exhibit a high degree of similarity, indicating substantial redundancy in the global computation. 

Motivated by the observation of attention redundancy, we adapt the training-free technique of token merging~\cite{bolya2022token,bolya2023token,tran2024accelerating,lee2024video} to enhance VGGT's inference efficiency. Token merging consolidates redundant representations by partitioning tokens into source (src) and destination (dst) sets and merging each src token into its most similar dst counterpart. While effective in 2D vision tasks, its extension to structures designed for 3D geometry understanding remains underexplored. Unlike 2D settings that process single images, VGGT relies on cross-image correspondences, making direct application of token merging highly challenging. To address this, we introduce FastVGGT, a novel, training-free framework that strategically applies token merging to mitigate the Global Attention bottleneck. Our approach begins by preserving the foundational coordinate system. Specifically, tokens from the initial frame, which serves as the global reference for the entire scene, are designated as high-priority destination (dst) tokens and are exempt from being merged to ensure reconstruction stability. Furthermore, to maintain global consistency and preserve fine-grained details, we identify and retain the most salient tokens across all frames, allowing them to bypass the merging process entirely and participate directly in the attention computation. Finally, drawing inspiration from ToMeSD~\cite{bolya2023token}, we implement region-based random sampling within each subsequent frame. This ensures a spatially balanced selection of src and dst tokens, preventing critical information loss in localized regions during consolidation.

Our experiments demonstrate that this integrated approach allows FastVGGT to significantly reduce the computational overhead of Global Attention. For large-scale inputs of 1000 images, it achieves 4$\times$ inference speedup over the baseline VGGT while simultaneously mitigating error accumulation in long-sequence reconstructions. Notably, the original VGGT suffers from prohibitive memory consumption, leading to Out-of-Memory (OOM) errors when processing sequences beyond 300 images. Through VRAM optimization, our modified VGGT successfully handles inputs of over 1000 images, demonstrating a substantial improvement in scalability.

In summary, our main contributions are as follows: 1) We identify and analyze the key bottleneck that limits the inference speed of VGGT. 2) Based on our observation of VGGT’s Global Attention, we introduce token merging into feed-forward visual geometry architectures for the first time. 3) Extensive experiments demonstrate that our method significantly accelerates VGGT on large-scale inputs while preserving reconstruction quality and mitigating error accumulation.

%% file: sec/2_related.tex
\section{Related work}
\label{sec:formatting}
\subsection{Feed-Forward 3D Reconstruction}
Building on the foundations of traditional 3D reconstruction~\cite{mur2015orb,schonberger2016structure,wang2025look,shen2025evolving,mur2017orb}, recent end-to-end learning-based methods leverage neural networks to encode scene priors, substantially enhancing robustness and cross-dataset generalization. Early progress was marked by DUSt3R~\cite{wang2024dust3r}, which directly regresses view-consistent 3D point maps from only two RGB images without requiring camera calibration. Its successor, MASt3R~\cite{leroy2024grounding}, introduces confidence-weighted losses to approximate metric scale, further improving reconstruction quality. The current state-of-the-art, VGGT~\cite{wang2025vggt} scales this philosophy to a 1.2B-parameter transformer that jointly predicts camera intrinsics, extrinsics, dense depth, point maps, and 2D tracks. However, as input sequences grow longer, the global attention mechanism must capture inter-frame relations within an expanding token space. This not only increases computational overhead but also amplifies noise propagation, making long-sequence predictions more prone to drift. VGGT-Long~\cite{deng2025vggt} addresses the drifting issue by aligning sub-maps to suppress error accumulation, but at the cost of significantly reduced inference speed, undermining the efficiency of feed-forward 3D reconstruction. To overcome these challenges, we propose FastVGGT, which accelerates inference and mitigates error accumulation by reducing the number of tokens processed in Global Attention, thereby achieving a balance between efficiency and accuracy.

\subsection{Token Merging}
Visual token merging~\cite{bolya2022token,renggli2022learning,zeng2022not,haurum2023tokens} was first introduced as a training-free approach to improve the throughput of Vision Transformers (ViTs)~\cite{dosovitskiy2020image}. It was later extended to reduce computational cost in tasks such as diffusion, video and language understanding, and video editing~\cite{cao2023pumer,choi2024vid,shen2024tempme}. The method partitions tokens into two sets, matches each token with its most similar counterpart, merges them via average pooling, and concatenates the results. Although simple and effective, most applications remain limited to the image domain, while long-form video remains underexplored despite its spatiotemporal tokens being both redundant and interdependent. In parallel, other approaches to token reduction have emerged. TokenLearner~\cite{ryoo2021tokenlearner} adaptively selects informative tokens with an MLP. PuMer~\cite{cao2023pumer} integrates pruning and merging for vision–language models. Pooling-based methods accelerate attention by averaging token embeddings. While sharing the goal of efficient transformers, our work targets the unique demands of feed-forward 3D reconstruction, where tokens must retain spatial precision and temporal coherence. To this end, we propose FastVGGT, which adapts token merging to VGGT, achieving substantial acceleration while maintaining reconstruction fidelity.

%% file: sec/3_method.tex
\section{Method}
\label{sec:Method}

\subsection{Visualization Findings}
We highlight an observation that motivates our optimization of the Global Attention module. As shown in Figure~\ref{P3}, we visualize VGGT’s Global Attention maps on the ScanNet dataset. Each image is represented by 1,041 tokens (one camera token, four register tokens, and 1,036 patch tokens from a 28 $\times$ 37 grid). The dense self-attention mechanism generates an attention map for every token, and visualizations across tokens and blocks reveal that many of these maps are highly similar.

The phenomenon of attention similarity, often referred to as feature degradation, has been consistently reported in the DINO series~\cite{oquab2023dinov2,simeoni2025dinov3,caron2021emerging}. In DINO, while the CLS token becomes increasingly discriminative, patch-level features gradually lose local consistency as they converge toward the CLS token, which undermines performance on dense prediction tasks. A similar limitation appears in VGGT: its global self-attention layers aggregate information across all tokens via weighted averaging, and without explicit regularization or task-specific constraints, the representational space is repeatedly compressed. This compression drives tokens to collapse along a dominant direction, eroding their distinctiveness. At a fundamental level, both DINO and VGGT lack explicit mechanisms to preserve token diversity, leading to a progressive loss of local variation.

It is important to note that the consequences of feature degradation differ markedly between DINO and VGGT. In DINO, the network must handle not only image classification but also dense prediction tasks such as segmentation, where local feature integrity is critical. As training progresses, patch tokens collapse toward the CLS token, weakening dense feature discrimination and directly impairing downstream performance. This motivates corrective strategies such as Gram Anchoring~\cite{simeoni2025dinov3}. By contrast, VGGT follows a two-stage design: Global Attention and Frame Attention. Global Attention enforces global consistency by capturing holistic spatial–temporal relationships. Here, the apparent degradation—tokens collapsing toward a dominant subspace—can be interpreted not as a flaw but as deliberate distillation of global semantics. Frame Attention then reintroduces local variability, striking a balance between global abstraction and local differentiation. Moreover, the strong feature similarity observed in Global Attention exposes redundancy that can be exploited to reduce computational cost without compromising performance, providing a natural path to address VGGT’s speed bottleneck.

\begin{table}[t]
\small
	\tabcolsep=0.25cm
    \begin{center}
	\renewcommand{\arraystretch}{1.1}
	\begin{tabular}{c|cc|cc|c}
		\specialrule{1pt}{0pt}{2pt}   
		\specialrule{0.3pt}{0pt}{0pt} 
	\multirow{2}{*}{} & \multicolumn{2}{c|}{Random} & \multicolumn{2}{c|}{Fixed} & \multirow{2}{*}{Baseline} \\
	 & \textit{r} ~=~0.5 & \textit{r}~=~0.8 & \textit{s}~=~5 & \textit{s}~=~8 &  \\ \hline
	CD~$\downarrow$ & 0.32 & 0.44 & 0.33 & 0.77 & 0.21 \\
	Time~$\downarrow$ & 22.6 & 16.9 & 17.2 & 11.6 & 30.4 \\ \hline
	\end{tabular}\vspace{-1mm}
    \caption{Results of applying merging strategies based on random sampling and fixed-stride sampling. Here, \textit{r} denotes the merging ratio, and \textit{s} represents the stride for selecting dst tokens.}\label{t1}\vspace{-8mm}

    \end{center}
	\end{table}
    
Motivated by the strong similarities observed in Global Attention maps, we propose FastVGGT, which accelerates inference via token merging. Conventional strategies typically partition tokens into destination (dst) and source (src) sets through random or fixed-stride sampling. However, the structural and task-specific requirements of 3D reconstruction make their direct application to VGGT non-trivial. To assess their suitability, we tested both random sampling (with two merging ratios) and fixed-stride sampling (with two stride settings) in VGGT’s Global Attention layer. As shown in Table~\ref{t1}, these methods reduce inference time but significantly increase the Chamfer Distance, thereby degrading reconstruction accuracy and failing to retain VGGT’s strong performance. To overcome this, we design a tailored strategy for feed-forward 3D reconstruction that preserves accuracy, accelerates inference, and mitigates error accumulation in long-sequence scenarios.

\begin{figure*}[t]
	\centering
	\includegraphics[scale=0.53]{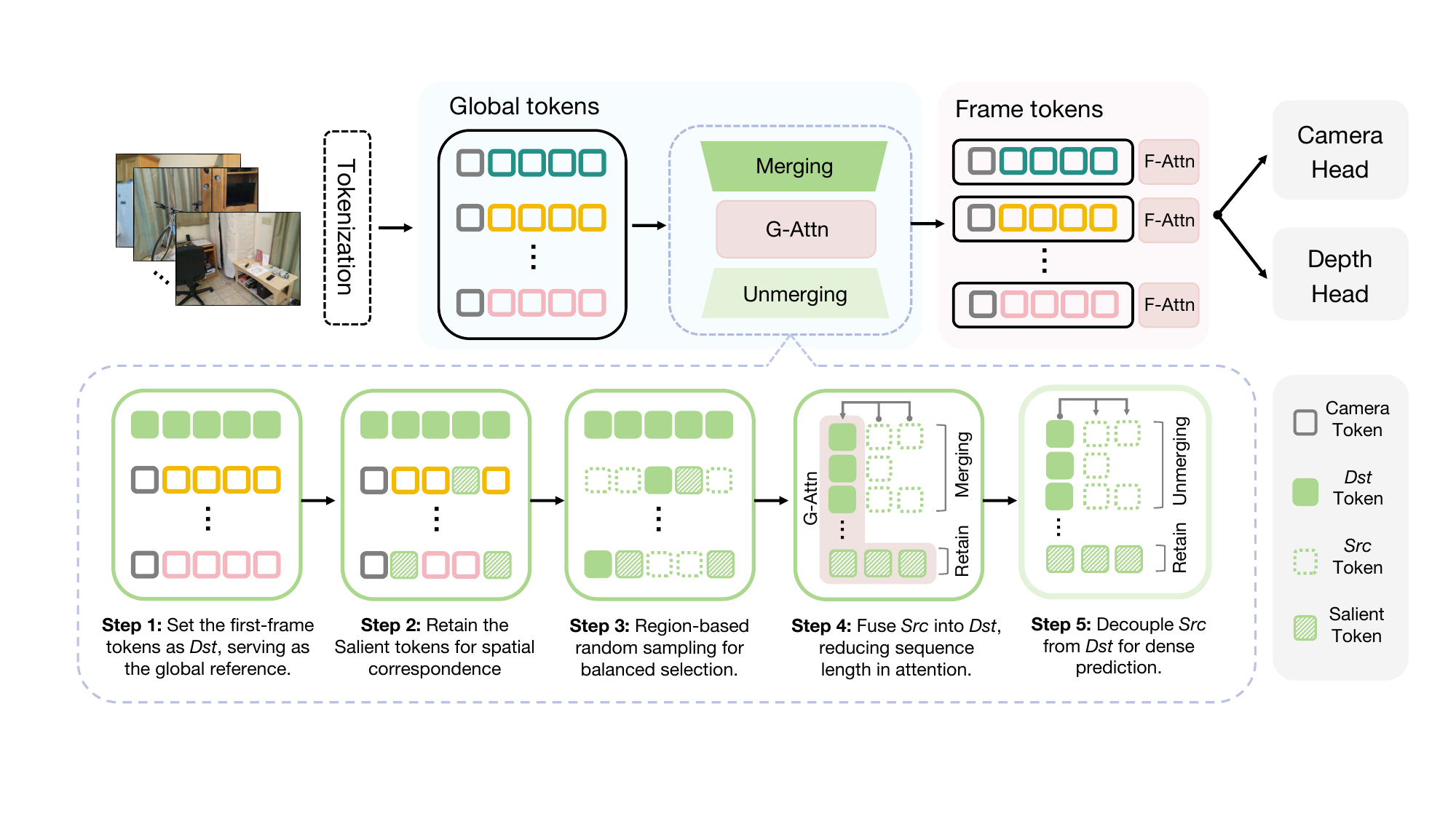}\vspace{0mm}
	\caption{Overview of the proposed token merging strategy. The pipeline begins with tokenization of input frames. To alleviate the Global Attention bottleneck, we design a five-step procedure: (1) fix initial-frame tokens as destination (\textit{Dst}) tokens, serving as the global reference for spatial consistency, (2) retain the top-k salient tokens to strengthen correspondences, (3) apply region-based random sampling for balanced selection, (4) fuse source (\textit{Src}) tokens into their nearest \textit{Dst} tokens during attention, and (5) decouple tokens via unmerging for dense reconstruction. G-Attn and F-Attn denote Global Attention and Frame Attention, respectively.}\label{P4}\vspace{0mm}
\end{figure*}

\subsection{Token Partitioning}
In contrast to token merging methods primarily developed for 2D vision tasks, which typically operate on individual images, feed-forward 3D reconstruction models process multi-view sequences with varying degrees of overlap across viewpoints. Such overlaps are critical for accurate 3D reconstruction, as they provide the cross-view correspondences necessary for recovering consistent geometry. Therefore, to achieve acceleration through merging while preserving reconstruction fidelity, it becomes essential to design an effective token partitioning strategy that operates across all frames in the sequence.

The design of our token partitioning strategy is governed by three principles: 1) preserving cross-frame correspondence to ensure structural consistency of the overall scene; 2) ensuring uniform merging within each frame to prevent both over-compression and redundancy; and 3) merging the most redundant tokens into the most representative ones to maximize acceleration efficiency. Accordingly, we divide tokens into three categories: salient tokens, which capture the most distinctive features of each frame; destination (dst) tokens, which act as representative anchors; and source (src) tokens, which correspond to redundant information to be merged. Based on these principles, we design three token partitioning strategies as detailed in the following.

\noindent \textbf{Reference Token Selection.} First, VGGT defines the first frame as the world coordinate system, with all tokens registered relative to this reference. This design makes the first frame a key anchor for maintaining spatial consistency across the sequence. Supplementary visualization of attention maps further shows that tokens consistently exhibit stronger activations toward the first frame than any other, highlighting its central role in guiding scene-level representations. Consequently, we designate all tokens from the first frame as dst tokens due to their strong representativeness.

\noindent \textbf{Salient Token Selection.} Second, since VGGT reconstructs scenes through cross-frame token interactions, a subset of key tokens is critical for establishing reliable correspondences across views. As illustrated in Figure~\ref{P3}, these tokens resemble distinctive keypoints in traditional matching algorithms. To preserve them, we extend conventional token merging by introducing a third category, dividing tokens into salient, dst, and src groups. Salient tokens are excluded from merging operations and instead participate directly in attention. For selecting salient tokens, we first apply a top-k strategy based on token norms to measure distinctiveness. However, its computational overhead grows with longer sequences. To improve efficiency, we adopt a fixed-stride sampling scheme that retains about 10\% of tokens per frame as salient tokens, serving as anchors to maintain distinctiveness. Experiments show that this achieves accuracy comparable to top-k selection while greatly reducing cost. We therefore use fixed-stride sampling as the default strategy, balancing efficiency and effectiveness.

\noindent \textbf{Uniform Token Sampling.} Finally, recognizing the dense prediction nature of 3D reconstruction, we ensure uniform intra-frame sampling to avoid local over-compression or redundancy. To this end, we assign dst and src tokens within each frame using a region-based random sampling strategy inspired by the success of ToMeSD~\cite{bolya2023token} in diffusion models. Concretely, we first partition the input tokens by frame and arrange them into a 2D grid of image patches. Within each grid cell, dst tokens are sampled according to a predefined merging ratio with stride $K$, while the remaining tokens are designated as src tokens. This region-based strategy ensures spatially balanced merging and prevents artifacts such as the disappearance of large areas. Consequently, the merging process is more coherent, and the reconstructed scene better preserves global structural stability.

\subsection{Token Merging Procedure}
After token partitioning, src tokens are merged into their most similar counterparts in dst, effectively reducing the sequence length for attention computation. Formally, given a token representation $x \in \mathbb{R}^c $ with feature dimension $c$, we compute cosine similarity between each $x_s \in \text{src} $ and all $x_d \in \text{dst}$:

\begin{equation}
    \text{sim}(x_s, x_d) = \frac{x_s \cdot x_d}{\|x_s\| \, \|x_d\|}.
\end{equation}

Each source token $x_s$ is assigned to its most similar destination token $x_d$, and updated by averaging:

\begin{equation}
    x_d’ = \frac{x_d + x_s}{2}.
\end{equation}

The updated $x_d’$ is retained while $x_s$ is temporarily discarded, thereby reducing the number of tokens processed in attention computations.

\subsection{Token Unmerging Procedure}
Dense 3D reconstruction requires per-token outputs (\textit{e.g.}, depth predictions). To satisfy this requirement, we adopt an unmerging operation, inspired by ToMeSD~\cite{bolya2023token}, which restores the original token resolution and maintains full compatibility with the VGGT architecture. Specifically, suppose two tokens $x_1, x_2 \in \mathbb{R}^c$ are merged into a single representation:

\begin{equation}
x_{1,2}^* = \frac{x_1 + x_2}{2}.
\end{equation}

During unmerging, this representation is replicated to recover the original sequence length:

\begin{equation}
x_1’ = x_{1,2}^* \quad x_2’ = x_{1,2}^*
\end{equation}

This replication guarantees that the token count matches the input resolution, allowing the decoder to produce dense outputs for every patch. Moreover, by maintaining an explicit src–dst mapping during merging, the unmerging process is both deterministic and efficient. This ensures that the model enjoys the computational benefits of a compressed sequence in the Global Attention layers, while still delivering the per-token predictions.

\subsection{VRAM-Efficient Implementation}
In our tests, the original VGGT encounters out-of-memory errors when processing sequences of around 300 frames. VGGT consists of 24 encoder blocks, but during inference only the outputs of layers 4, 11, 17, and 23 are required. Nevertheless, the original implementation stores intermediate results from all 24 blocks. To support longer input sequences, we introduce an simple optimized variant, VGGT$^*$, which discards unused intermediate outputs during inference. This reduces memory consumption and enables processing of up to 1000 frames without affecting reconstruction quality. Unless otherwise specified, all experiments in this paper use VGGT$^*$ as the baseline.\vspace{2mm}

%% file: sec/4_experiment.tex
\section{Experiments}
\subsection{Experimental Setup}
We evaluate FastVGGT on three benchmark datasets: ScanNet, NRGBD, and 7 Scenes. For ScanNet, which contains 1,500 scenes, we uniformly sample 50 scenes to form a reproducible benchmark, denoted as ScanNet-50. Our experiments focus on two tasks: camera pose estimation and point map reconstruction. Across both tasks, FastVGGT consistently achieves substantial speedups while maintaining accuracy. The overall architecture follows VGGT, comprising L=24 frame and global attention layers, with Flash-Attention2~\cite{dao2023flashattention} integrated to further accelerate inference. All experiments are conducted on a workstation with an NVIDIA A800 GPU (80 GiB VRAM).\vspace{2mm}

	\begin{table*}[t]
		\small
		\centering
		\tabcolsep=0.46cm
		\renewcommand{\arraystretch}{1.3}
		\begin{tabular}{c|cc|cc|cc|cc}
			\specialrule{1pt}{0pt}{2pt}   
			\specialrule{0.3pt}{0pt}{0pt} 
			\multirow{2}{*}{Method} & \multicolumn{2}{c|}{1000} & \multicolumn{2}{c|}{500} & \multicolumn{2}{c|}{300} & \multicolumn{2}{c}{100} \\ \cline{2-9} 
			 & CD~$\downarrow$ & Time~$\downarrow$ & CD~$\downarrow$ & Time~$\downarrow$ & CD~$\downarrow$ & Time~$\downarrow$ & CD~$\downarrow$ & Time~$\downarrow$ \\ \hline
			 $\pi ^{3}$~\cite{wang2025pi} & \textit{\textit{OOM}} & \textit{OOM} & \textit{OOM} & \textit{OOM} & \textit{OOM} & \textit{OOM} & \textit{OOM} & \textit{OOM} \\
			 StreamVGGT\cite{zhuo2025streaming} & \textit{OOM} & \textit{OOM} & \textit{OOM} & \textit{OOM} & \textit{OOM} & \textit{OOM} & \textit{OOM} & \textit{OOM} \\
			 Fast3R~\cite{yang2025fast3r} & 0.684 & 397.8s & 0.701 & 97.3s & 0.711 & 34.9s & 0.723 & 4.8s \\
			 CUT3R~\cite{wang2025continuous} & 0.786 & 34.8s & 0.774 & 18.8s & 0.775 & 11.1s & 0.767 & 3.6s \\
			 VGGT$^*$~\cite{wang2025vggt} & 0.471 & 724.6s & 0.420 & 177.5s & 0.416 & 131.4s & 0.423 & 9.1s \\
             \rowcolor[HTML]{E4F9F9}
			 FastVGGT & \textbf{0.425} & \textbf{180.7s} & \textbf{0.411} & \textbf{55.2s} & \textbf{0.416} & \textbf{23.8s} & \textbf{0.426} & \textbf{5.4s} \\
			\hline
		\end{tabular}\vspace{0mm}\caption{Quantitative results of point cloud reconstruction on the ScanNet-50 dataset with input sequences of 1000, 500, 300, and 100 images. \textit{OOM} denotes out-of-memory.}\label{t2}
	\end{table*}
    
\begin{table*}[t]
	\small
	\centering
	\tabcolsep=0.15cm
\renewcommand{\arraystretch}{1.5}
	\begin{tabular}{c|ccccccc|ccccccc}
		\specialrule{1pt}{0pt}{2pt}   
		\specialrule{0.3pt}{0pt}{0pt} 
		\multirow{3}{*}{Method} & \multicolumn{7}{c|}{7 Scenes - Stride 3} & \multicolumn{7}{c}{7 Scenes - Stride 10} \\ \cline{2-15} 
		& \multicolumn{2}{c}{Acc~$\downarrow$} & \multicolumn{2}{c}{Comp~$\downarrow$} & \multicolumn{2}{c}{NC~$\uparrow$} & \multirow{2}{*}{Time $\downarrow$} & \multicolumn{2}{c}{Acc~$\downarrow$} & \multicolumn{2}{c}{Comp~$\downarrow$} & \multicolumn{2}{c}{NC~$\uparrow$} & \multirow{2}{*}{Time $\downarrow$} \\ \cline{2-7} \cline{9-14}
		& Mean & Med. & Mean & Med. & Mean & Med. &  & Mean & Med. & Mean & Med. & Mean & Med. &  \\ \hline
		$\pi ^{3}$~\cite{wang2025pi} & \textit{OOM} & \textit{OOM} & \textit{OOM} & \textit{OOM} & \textit{OOM} & \textit{OOM} & \textit{OOM} & \textit{OOM} & \textit{OOM} & \textit{OOM} & \textit{OOM} & \textit{OOM} & \textit{OOM} & \textit{OOM} \\
		StreamVGGT~\cite{zhuo2025streaming} & \textit{OOM} & \textit{OOM} & \textit{OOM} & \textit{OOM} & \textit{OOM} & \textit{OOM} & \textit{OOM} & \textit{OOM} & \textit{OOM} & \textit{OOM} & \textit{OOM} & \textit{OOM} & \textit{OOM} & \textit{OOM} \\  
		Fast3R~\cite{yang2025fast3r} & 0.045 & 0.027 & 0.047 & 0.010 & 0.616 & 0.627 & 43.7s & 0.040 & 0.021 & 0.056 & 0.013 & 0.639 & 0.657 & 5.5s \\
	   CUT3R~\cite{wang2025continuous} & 0.179 & 0.121 & 0.097 & 0.043 & 0.588 & 0.581 & 14.5s & 0.041 & 0.021 & 0.029 & 0.010 & 0.651 & 0.677 & 4.2s \\
	   VGGT$^*$~\cite{wang2025vggt} & 0.019 & 0.008 & 0.027 & 0.010 & 0.611 & 0.628 & 76.7s & 0.020 & 0.008 & 0.027 & 0.010 & 0.623 & 0.641 & 8.7s \\  \rowcolor[HTML]{E4F9F9}
	   FastVGGT &  \textbf{0.018} & \textbf{0.008} & \textbf{0.026} & \textbf{0.010} & \textbf{0.617} & \textbf{0.634} & \textbf{28.0s} & \textbf{0.018} & \textbf{0.008} & \textbf{0.027} & \textbf{0.010} & \textbf{0.628} & \textbf{0.648} & \textbf{5.1s} \\ \hline
	   \end{tabular}\vspace{0mm}
       	\caption{Quantitative results of point cloud reconstruction on the 7 Scenes dataset. Stride denotes keyframes sampled every 3 or 10 frames.}\label{t3}\vspace{0mm}
	\end{table*}

	\begin{table*}[t]
		\small
		\centering
		\tabcolsep=0.15cm
	\renewcommand{\arraystretch}{1.4}
		\begin{tabular}{c|ccccccc|ccccccc}
			\specialrule{1pt}{0pt}{2pt}   
			\specialrule{0.3pt}{0pt}{0pt} 
			\multirow{3}{*}{Method} & \multicolumn{7}{c|}{NRGBD - Stride 3} & \multicolumn{7}{c}{NRGBD - Stride 10} \\ \cline{2-15} 
			& \multicolumn{2}{c}{Acc~$\downarrow$} & \multicolumn{2}{c}{Comp~$\downarrow$} & \multicolumn{2}{c}{NC~$\uparrow$} & \multirow{2}{*}{Time $\downarrow$} & \multicolumn{2}{c}{Acc~$\downarrow$} & \multicolumn{2}{c}{Comp~$\downarrow$} & \multicolumn{2}{c}{NC~$\uparrow$} & \multirow{2}{*}{Time $\downarrow$} \\ \cline{2-7} \cline{9-14}
			& Mean & Med. & Mean & Med. & Mean & Med. &  & Mean & Med. & Mean & Med. & Mean & Med. &  \\ \hline
			$\pi ^{3}$~\cite{wang2025pi} & \textit{OOM} & \textit{OOM} & \textit{OOM} & \textit{OOM} & \textit{OOM} & \textit{OOM} & \textit{OOM} & \textit{OOM} & \textit{OOM} & \textit{OOM} & \textit{OOM} & \textit{OOM} & \textit{OOM} & \textit{OOM} \\
			StreamVGGT~\cite{zhuo2025streaming} & \textit{OOM} & \textit{OOM} & \textit{OOM} & \textit{OOM} & \textit{OOM} & \textit{OOM} & \textit{OOM} & \textit{OOM} & \textit{OOM} & \textit{OOM} & \textit{OOM} & \textit{OOM} & \textit{OOM} & \textit{OOM} \\
			Fast3R~\cite{yang2025fast3r} & 0.074 & 0.036 & 0.024 & 0.011 & 0.658 & 0.682 & 68.9s & 0.061 & 0.028 & 0.031 & 0.013 & 0.669 & 0.712 & 7.4s \\
		   CUT3R~\cite{wang2025continuous} & 0.346 & 0.243 & 0.184 & 0.090 & 0.579 & 0.623 & 18.3s & 0.132 & 0.064 & 0.056 & 0.0106 & 0.669 & 0.725 & 5.7s \\
		   VGGT$^*$~\cite{wang2025vggt} & 0.029 & 0.019 & 0.018 & 0.009 & 0.727 & 0.728 & 136.1s & 0.016 & 0.010 & 0.017 & 0.009 & 0.735 & 0.738 & 13.9s \\
           \rowcolor[HTML]{E4F9F9}
		   FastVGGT & \textbf{0.029} & \textbf{0.021} & \textbf{0.019} & \textbf{0.010} & \textbf{0.730} & \textbf{0.738} & \textbf{53.1s} 
& \textbf{0.018} & \textbf{0.011} & \textbf{0.018} & \textbf{0.009} & \textbf{0.736} & \textbf{0.741} & \textbf{7.3s} \\ \hline 
		   \end{tabular}\vspace{0mm}		\caption{Quantitative results of point cloud reconstruction on the NRGBD dataset.}\label{t4}\vspace{2mm}
		\end{table*}

	\begin{table*}[t]
		\small
		\centering
		\tabcolsep=0.48cm
		\renewcommand{\arraystretch}{1.5}
		\begin{tabular}{c|cc|cc|cc|cc}
			\specialrule{1pt}{0pt}{2pt}   
			\specialrule{0.3pt}{0pt}{0pt} 
\multirow{2}{*}{\begin{tabular}[c]{@{}c@{}}Input\\ Frames\end{tabular}} & \multicolumn{2}{c|}{ATE~$\downarrow$} & \multicolumn{2}{c|}{ARE~$\downarrow$} & \multicolumn{2}{c|}{RPE-rot~$\downarrow$} & \multicolumn{2}{c}{RPE-trans~$\downarrow$} \\ \cline{2-9} 
& Baseline & Ours & Baseline & Ours & Baseline & Ours & Baseline & Ours \\ \hline
1000 & 0.196 & \cellcolor[HTML]{E4F9F9}\textbf{0.164} & 4.636 & \cellcolor[HTML]{E4F9F9}\textbf{3.860} & 0.997 & \cellcolor[HTML]{E4F9F9}\textbf{0.667} & 0.039 & \cellcolor[HTML]{E4F9F9}\textbf{0.029} \\
500  & 0.174 & \cellcolor[HTML]{E4F9F9}\textbf{0.145} & 4.190 & \cellcolor[HTML]{E4F9F9}\textbf{3.591} & 0.963 & \cellcolor[HTML]{E4F9F9}\textbf{0.627} & 0.042 & \cellcolor[HTML]{E4F9F9}\textbf{0.031} \\
300  & 0.145 & \cellcolor[HTML]{E4F9F9}\textbf{0.142} & 3.689 & \cellcolor[HTML]{E4F9F9}\textbf{3.554} & 0.786 & \cellcolor[HTML]{E4F9F9}\textbf{0.801} & 0.040 & \cellcolor[HTML]{E4F9F9}\textbf{0.036} \\
100  & 0.140 & \cellcolor[HTML]{E4F9F9}\textbf{0.141} & 3.625 & \cellcolor[HTML]{E4F9F9}\textbf{3.512} & 1.224 & \cellcolor[HTML]{E4F9F9}\textbf{1.262} & 0.058 & \cellcolor[HTML]{E4F9F9}\textbf{0.061} \\ \hline
		\end{tabular}\vspace{0mm}\caption{Quantitative results of camera pose estimation on the ScanNet-50 dataset.}\label{t5}\vspace{0mm}
	\end{table*}

\subsection{3D Reconstruction}
\textbf{Comparisons on the ScanNet-50 Dataset.
} We begin by evaluating FastVGGT on the ScanNet-50 dataset, reporting reconstruction quality using Chamfer Distance (CD). Experiments are conducted with input sequences of 1000, 500, and 100 images, enabling us to assess performance under varying sequence lengths. The results in Table~\ref{t2} show that although SOTA methods such as $\pi^3$ and StreamVGGT achieve strong performance on short sequences, they fail on long sequences due to memory constraints. Methods like Fast3R~\cite{yang2025fast3r} and CUT3R~\cite{wang2025continuous} can process long sequences efficiently, but their reconstruction quality degrades severely. In contrast, FastVGGT delivers substantial acceleration over baseline VGGT across all settings while preserving reconstruction accuracy. Notably, when processing very long sequences ($e.g.$, 1000 images), FastVGGT not only maintains reconstruction fidelity but also significantly mitigates error accumulation, demonstrating robustness and scalability for large-scale 3D reconstruction.

\noindent \textbf{Comparisons on 7 Scenes and NRGBD Datasets.}
Following the CUT3R protocol, we evaluate FastVGGT on the 7 Scenes and NRGBD datasets. We report accuracy (Acc), completeness (Comp), and normal consistency (NC) using long-sequence inputs, with keyframes sampled every 3 or 10 frames. As shown in Table~\ref{t3} and Table~\ref{t4}, FastVGGT maintains the robust performance demonstrated on ScanNet-50, further demonstrating its effectiveness for 3D reconstruction.

\begin{figure}[t]
	\centering 
	\includegraphics[scale=0.58]{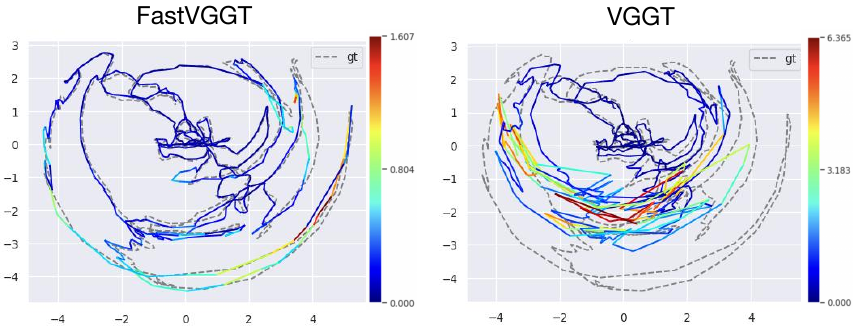}
	\caption{Comparison of pose estimation performance between FastVGGT and VGGT.
}\label{P5}\vspace{0mm}
\end{figure}
\subsection{Camera Pose Estimation}
We evaluate FastVGGT on the ScanNet-50 dataset for camera pose estimation. Evaluation is conducted using four commonly adopted metrics: Absolute Trajectory Error (ATE), reflecting trajectory-level accuracy; Absolute Rotation Error (ARE), reflecting orientation accuracy; Relative Pose Error in rotation (RPE-rot), and Relative Pose Error in translation (RPE-trans), which jointly characterize local frame-to-frame consistency. As illustrated in Table~\ref{t5}, FastVGGT matches the baseline VGGT on shorter sequences (100 and 300 frames), while on longer sequences (500 and 1000 frames) it substantially reduces pose estimation error. Figure~\ref{P5} further visualizes pose trajectories, demonstrating FastVGGT’s ability to suppress cumulative drift, thereby underscoring its effectiveness and practicality for real-world deployment.

\begin{table}[]
\begin{center}
\small
\tabcolsep=0.18cm
\renewcommand{\arraystretch}{1.3}
\begin{tabular}{c|ccc|c|c}
\Xhline{1px}\hline
Methods & Uniform & Reference  & Salient  & CD~$\downarrow$ & ATE~$\downarrow$ \\ \hline
(a) & - & - & - & 0.947 & 0.842 \\
(b) & \checkmark & - & - & 0.637 & 0.722 \\
(c) & \checkmark & \checkmark & - & 0.431 & 0.149 \\
(d) & \checkmark & \checkmark & \checkmark & 0.411 & 0.141 \\ \hline
\end{tabular}
\caption{Ablation study for Token partitioning strategies.}\label{t6}
\end{center}\vspace{0mm}
\end{table}

\begin{table}[]
\begin{center}
\small
\tabcolsep=0.14cm
\renewcommand{\arraystretch}{1.3}
\begin{tabular}{c|cc|cc|cc}
\Xhline{1px}\hline
\multirow{2}{*}{Blocks} & \multicolumn{2}{c|}{0.3} & \multicolumn{2}{c|}{0.6} & \multicolumn{2}{c}{0.9} \\ \cline{2-7} 
 & CD$~\downarrow$ & Time$~\downarrow$ & CD$~\downarrow$ & Time$~\downarrow$ & CD$~\downarrow$ & Time$~\downarrow$ \\ \hline
0 & 0.408 & 119.8s & 0.415 & 64.3s & 0.411 & 55.2s \\
10 & 0.418 & 146.2s & 0.424 & 118.4s & 0.431 & 106.3s \\
20 & 0.423 & 172.9s & 0.427 & 169.1s & 0.411 & 157.3s \\ \hline
\end{tabular}
\caption{Ablation study on merging ratio and affected blocks.}\label{t7}
\end{center}\vspace{0mm}
\end{table}

\subsection{Ablation Studies}
\textbf{Token Partitioning.} We evaluate the effectiveness of different token partitioning strategies using 500-frame inputs from ScanNet-50. As shown in Table~\ref{t6}: (a) directly selecting dst and src tokens through random sampling yields poor performance; (b) adopting region-based intra-frame uniform sampling improves performance but remains suboptimal; (c) designating the global reference frame (first frame) as dst tokens brings substantial gains; and (d) further protecting salient tokens leads to the best performance.

\noindent \textbf{Location and Intensity of Merging.} To balance accuracy and efficiency, we evaluate the effect of varying the starting block for merging and the applied merging ratio. As shown in Table~\ref{t7}, increasing the merging ratio consistently reduces inference time, with only minor fluctuations in Chamfer Distance. Consequently, we adopt an aggressive strategy that applies a 90\% merging ratio from block 0 across all subsequent layers, yielding a favorable balance of accuracy and efficiency.

\section{Conclusion}
In this work, we proposed FastVGGT, a training-free method that accelerates VGGT inference via strategic token merging without sacrificing reconstruction quality. By profiling VGGT, we identified the global attention module as the primary bottleneck for long-sequence inputs and observed strong similarities in its attention maps. To address this, we introduce token merging to alleviate the inference bottleneck and design strategies tailored to the feed-forward visual geometry model. Experiments on multiple benchmarks show that FastVGGT achieves up to 4x speedup on 1000-images inputs while maintaining competitive accuracy in both camera pose estimation and 3D reconstruction tasks, and further mitigates error accumulation in long-sequence settings. These results highlight token merging as a principled solution for scaling visual geometry models and demonstrate FastVGGT's immediate practicality for real-world applications.